\documentclass[preprint,5p,twocolumn,12pt]{elsarticle}

\usepackage{amssymb}
\usepackage{amsmath}

\usepackage[T1]{fontenc}
\usepackage{booktabs}
\usepackage{multirow}
\usepackage[nohyperlinks, nolist]{acronym}
\usepackage{xspace}
\usepackage{url}
\usepackage[hidelinks]{hyperref}

\journal{Energy and AI}

\begin{document}

\begin{frontmatter}



\title{Learning the Optimal Power Flow: Environment Design Matters}


\author[inst1]{Thomas Wolgast\corref{cor1}}
\cortext[cor1]{Corresponding Author}
\ead{thomas.wolgast@uni-oldenburg.de}
\author[inst1]{Astrid Nieße}
\ead{astrid.niesse@uni-oldenburg.de}

\affiliation[inst1]{organization={Carl von Ossietzky Universität Oldenburg, Department of Computing Science,  Digitalized Energy Systems},
            addressline={Ammerländer Heerstraße 114-118}, 
            city={Oldenburg},
            postcode={26129}, 
            country={Germany}}

\begin{abstract}
To solve the optimal power flow (OPF) problem, reinforcement learning (RL) emerges as a promising new approach. However, the RL-OPF literature is strongly divided regarding the exact formulation of the OPF problem as an RL environment. In this work, we collect and implement diverse environment design decisions from the literature regarding training data, observation space, episode definition, and reward function choice. In an experimental analysis, we show the significant impact of these environment design options on RL-OPF training performance. Further, we derive some first recommendations regarding the choice of these design decisions. The created environment framework is fully open-source and can serve as a benchmark for future research in the RL-OPF field.
\end{abstract}




\end{frontmatter}

\newcommand{\eg}{e.g.\xspace}
\newcommand{\ie}{i.e.\xspace}
\newcommand{\etal}{et al.\xspace}

\newcommand{\pandapower}{\textit{pandapower}\xspace}
\newcommand{\simbench}{\textit{SimBench}\xspace}
\newcommand{\gymnasium}{\textit{Gymnasium}\xspace}
\newcommand{\mujoko}{\textit{MuJoCo}\xspace}
\newcommand{\pybullet}{\textit{PyBullet}\xspace}

\newcommand{\reset}{\textit{reset}\xspace}
\newcommand{\step}{\textit{step}\xspace}

\newcommand{\blind}{\emph{removed for blindness}}

\newcommand{\timeseries}{\textit{Time-Series}\xspace}
\newcommand{\normalSampling}{\textit{Normal}\xspace}
\newcommand{\uniformSampling}{\textit{Uniform}\xspace}

\newcommand{\markov}{\textit{Markov}\xspace}
\newcommand{\redundantObs}{\textit{Redundant}\xspace}
\newcommand{\removeObs}{\textit{Remove}\xspace}

\newcommand{\onestep}{\textit{1-Step}\xspace}
\newcommand{\nstep}{\textit{n-Step}\xspace}

\newcommand{\summation}{\textit{Summation}\xspace}
\newcommand{\replacement}{\textit{Replacement}\xspace}

\newcommand{\timeseriesShort}{\textit{Time-Series}\xspace}
\newcommand{\normalSamplingShort}{\textit{Normal}\xspace}
\newcommand{\uniformSamplingShort}{\textit{Uniform}\xspace}

\newcommand{\markovShort}{\textit{Markov}\xspace}
\newcommand{\redundantObsShort}{\textit{Redundant}\xspace}
\newcommand{\removeObsShort}{\textit{Remove}\xspace}

\newcommand{\onestepShort}{\textit{1-Step}\xspace}
\newcommand{\nstepShort}{\textit{n-Step}\xspace}

\newcommand{\summationShort}{\textit{Summation}\xspace}
\newcommand{\replacementShort}{\textit{Replacement}\xspace}

\newcommand{\voltagecontrol}{VoltageControl\xspace}
\newcommand{\ecodis}{EcoDispatch\xspace}

\newcommand{\pandapowerfootnote}{\footnote{\url{https://pandapower.readthedocs.io/en/latest/}, last access: 2023-10-30}\xspace}
\newcommand{\simbenchfootnote}{\footnote{\url{https://simbench.de/en/}, last access: 2023-10-30}\xspace}
\newcommand{\gymnasiumfootnote}{\footnote{\url{https://gymnasium.farama.org/index.html}, last access: 2024-01-10}\xspace}
\newcommand{\mujokofootnote}{\footnote{\url{https://gymnasium.farama.org/environments/mujoco/}, last access: 2024-01-17}\xspace}
\newcommand{\footnotePybullet}{\footnote{\url{https://pybullet.org}, last access: 2024-01-25}\xspace}

\newcommand{\minSub}{\mathrm{min}}
\newcommand{\maxSub}{\mathrm{max}}
\newcommand{\activepower}{P}
\newcommand{\reactivepower}{Q}
\newcommand{\apparentPower}{S}
\newcommand{\objective}{J}
\newcommand{\objectiveFct}{J(s)}
\newcommand{\penaltyFct}{P(s)}
\newcommand{\violation}{v}
\newcommand{\penaltyFactor}{\alpha}
\newcommand{\constraints}{C}
\newcommand{\constraint}{c}
\newcommand{\constant}{k}
\newcommand{\generator}{g}
\newcommand{\generators}{G}
\newcommand{\load}{}
\newcommand{\Min}{\mathrm{min}\;}
\newcommand{\price}{p}
\newcommand{\voltage}{U}
\newcommand{\bus}{b}
\newcommand{\buses}{B}
\newcommand{\lineSingle}{l}
\newcommand{\lines}{L}
\newcommand{\loading}{S}
\newcommand{\grid}{\mathrm{ext}}
\newcommand{\forallGens}{\; \forall \; \generator \in \generators}
\newcommand{\maxReactive}{\reactivepower_{\maxSub, \generator}}
\newcommand{\act}{a}
\newcommand{\obs}{o}
\newcommand{\valid}{\mathrm{valid}}
\newcommand{\invalidShare}{\psi}

\newcommand{\nScatter}{1000\xspace}
\newcommand{\nInit}{1000\xspace}
\newcommand{\nTest}{6720\xspace}
\newcommand{\nRuns}{10\xspace}
\newcommand{\nCategories}{4\xspace}
\newcommand{\nOptions}{10\xspace}
\newcommand{\nVariants}{10\xspace}

\begin{acronym}[JSONP]\itemsep0pt
	\acro{DER}{distributed energy resource}
	\acro{DSO}{distribution system operator}
	\acro{EHV}{extra high voltage}
	\acro{HV}{high voltage}
	\acro{LV}{low voltage}
	\acro{MV}{medium voltage}
	\acro{OPF}{optimal power flow}
	\acro{PV}{photovoltaic}
	\acro{RES}{renewable energy resources}
	\acro{TSO}{transmission system operator}
	\acro{WT}{wind turbine}
	\acro{RL}{reinforcement learning}
	\acro{DRL}{deep RL}
	\acro{MARL}{multi-agent \acl{RL}}
	\acro{NE}{Nash-equilibrium}
	\acro{MAPE}{mean absolute percentage error}
	
	\acro{NN}{artificial neural network}
	\acro{ML}{machine learning}
	\acro{MADDPG}{Multi-Agent Deep Deterministic Policy Gradient}
	\acro{MMADDPG}[M-MADDPG]{model-extended MADDPG}
    \acro{MATD3}{Multi-Agent Twin-Delayed \ac{DDPG}}
	\acro{DDPG}{Deep Deterministic Policy Gradient}
	\acro{DDQN}{Double Deep Q Network}
	\acro{TD3}{Twin-Delayed \ac{DDPG}}
	\acro{MDP}{Markov decision process}
	\acro{WoLF-PHC}{win or learn fast policy hill climbing}

\end{acronym}


\section{Introduction}
The \ac{OPF} is a broad class of optimization problems in the energy field \cite{frankOptimalPowerFlow2012}. The common denominator is to optimize the power flow in a given power system to minimize an objective function subject to constraints. The current transformation of the energy system places new demands on the \ac{OPF}.
To achieve a real-time and continuous optimization of electric power systems, the \ac{OPF} needs to be performed more frequently and potentially for various scenarios. That requires a fast and real-time capable solver \cite{zamzamLearningOptimalSolutions2020a, yanRealTimeOptimalPower2020}. If researchers use the \ac{OPF} to investigate diverse scenarios under different influencing factors with millions of variants \cite{wolgastReinforcementLearningVulnerability2021}, the speed-solving requirements increase even more.
Conventionally, the \ac{OPF} is solved with methods like interior point or Newton's method \cite{frankOptimalPowerFlow2012}, which are too slow when thousands or millions of \ac{OPF} solutions are required in a high frequency \cite{zamzamLearningOptimalSolutions2020a, yanRealTimeOptimalPower2020}. 
\par
To speed up computation, a common approach of recent years is to train Deep Neural Networks to approximate the \ac{OPF} \cite{panDeepOPFFeasibilityOptimizedDeep2022, zhouDeepOPFFTOneDeep2023, zhouDatadrivenMethodFast2020a,nieDeepReinforcementLearning2022,panDeepOPFALAugmentedLearning2023}. This way, at the cost of a computationally heavy training phase, the difficult optimization problem is translated to a series of matrix multiplications, resulting in a speed-up of several orders of magnitudes.  
The training can be done by using supervised learning \cite{zamzamLearningOptimalSolutions2020a,panDeepOPFFeasibilityOptimizedDeep2022, zhouDeepOPFFTOneDeep2023,
panDeepOPFALAugmentedLearning2023}, \ie, learning by examples of existing optimal solutions, or by using \ac{RL} \cite{yanRealTimeOptimalPower2020, sayedFeasibilityConstrainedOnline2022, wooRealTimeOptimalPower2020,zhouDeepReinforcementLearning2022}, \ie, solving the problem by trial-and-error without pre-existing optimal solutions. Both options show promising results and are under intense research. They also can be combined by using supervised learning for pre-training and \ac{RL} for fine-tuning \cite{zhouDatadrivenMethodFast2020a,zhouDeepReinforcementLearning2022}. However, \ac{RL} has the big advantage of not requiring a conventional solver for training data generation \cite{sayedFeasibilityConstrainedOnline2022}.
\par 
In \ac{RL}, an \textit{agent} interacts with an \textit{environment}, which represents the problem to solve, which is the \ac{OPF} in this case. The agent performs sequential actions based on its observations of the environment's state and receives a reward for each action. The goal is to maximize the sum of rewards \cite{suttonReinforcementLearningSecond2018}. 
\par 
Most \ac{RL} publications focus on the \ac{RL} algorithm and how training can be improved for given \ac{RL} benchmark problems, \eg, the \mujoko\mujokofootnote environments. To serve as benchmarks for \ac{RL} research, these environments should be designed to be difficult but possible to solve \cite{kurachGoogleResearchFootball2020}. 
In contrast to that, when applying \ac{RL} to practical problems, we should design the simplest environment possible for a given problem, \ie, supporting the training process instead of challenging it. However, the literature on \ac{RL} environment design for best training performance is very limited \cite{redaLearningLocomoteUnderstanding2020}. Reda \etal \cite{redaLearningLocomoteUnderstanding2020} systematically compare different design options in the \pybullet\footnotePybullet locomotion environments regarding their influence on training performance. They demonstrate that environment design can significantly influence \ac{RL} performance while being heavily neglected by researchers at the same time. 
\par 
In line with the lack of literature regarding environment design, most \ac{RL}-\ac{OPF} publications choose very different environment designs when formalizing the \ac{OPF} problem as \ac{RL} environment, suggesting a lack of consensus in the literature about how the \ac{OPF} should be formulated as  \ac{RL} environment (see section \ref{sec:related}). Following the findings from Reda \etal \cite{redaLearningLocomoteUnderstanding2020}, we hypothesize that environment design can significantly influence \ac{RL}-\ac{OPF} performance and should be actively researched. 
In this work, we focus on how \ac{OPF} environments need to be designed to support the training process of \ac{RL} agents, regarding both optimization and constraint satisfaction performance of the \ac{OPF}.
\par 
\par 
Our contributions are as follows:
\begin{enumerate}
    \item We identify and describe four different design decision categories from the \ac{RL}-\ac{OPF} literature with multiple options within each category.
    \item We implement 13 different design variants overall and compare their performance after training in two different \ac{OPF} problems.
    \item With our experiments, we show the significant impact of these design decisions on training performance.
    \item From our results, we derive first insights and recommendations regarding design of \ac{RL}-\ac{OPF} environments for future research. 
    \item We open-source our developed \ac{OPF} environment framework and all source code for other researchers.\footnote{\url{https://github.com/Digitalized-Energy-Systems/rl-opf-env-design}}
\end{enumerate}
With this, our overall contribution is a guideline on how to evaluate relevant design options for \ac{RL}-\ac{OPF} problems.
\par 
Our work is structured as follows: In the following section \ref{sec:problem}, we first identify the special challenges of the \ac{OPF} when formulated as \ac{RL} problem and derive some limiting assumptions for our work. Section \ref{sec:openDesign} then identifies various design options from the \ac{RL}-\ac{OPF} literature and discusses the underlying idea as well as expected advantages and disadvantages of the respective options. Section \ref{sec:framework} presents our \ac{OPF} environment framework and section \ref{sec:experimentation} presents the experimentation details, including the two exemplary \ac{OPF} problems used in this work. Section \ref{sec:results} evaluates our results and provides recommendations for each category of design options respectively, followed by some concluding remarks.

\section{Challenges of the OPF as RL Problem}\label{sec:problem}
Before looking into the environment design, we will first discuss the special characteristics of the \ac{RL}-\ac{OPF} problem that make it a challenging \ac{RL} problem. These challenges are constraint satisfaction, large action spaces, high computational effort for evaluation, and training data. 
\par 
First, the \ac{OPF} problem is usually a constrained optimization problem. However, constraints are not part of the standard \ac{RL} framework. Therefore, either special \ac{RL} algorithms are required, or the reward function must include a penalty function that enables the agent to learn a valid policy \cite{khaloieReviewMachineLearning2024}.
Additionally, power systems are critical infrastructure. Therefore, in some cases, constraint violations are completely unacceptable. However, in other cases, minor violations are fully acceptable, \eg, slightly too high voltage magnitudes \cite{stottOptimalPowerFlow2012}. Whether violations can be tolerated depends on the exact use case, \eg, whether the model is used in simulation only or in actual grid operation.
\par 
Second, since the \ac{OPF} has mostly generator or load setpoints as degrees of freedom, the \ac{OPF} usually has a continuous action space. That limits the \ac{RL} algorithm choice, \eg, Deep Q-Learning \cite{mnihPlayingAtariDeep2013} is not applicable. Additionally, large-scale \ac{OPF} problems can have hundreds or thousands of degrees of freedom. In \ac{RL}, such high-dimensional action spaces are very challenging and not often investigated. For example, the popular \mujoko benchmark environments have only about six actions on average. 
\par 
Third, evaluating an agent's action, \ie, actuator setpoints, requires a power flow calculation to compute the resulting power flows and voltage values. This makes an \ac{OPF} environment computationally very expensive. Therefore, especially sample-efficient \ac{RL} algorithms should be used for training, which limits the \ac{RL} algorithm selection again.
\par
Fourth, realistic training data for the \ac{OPF} is very limited. In contrast to artificial environments like \mujoko environments, random sampling of states can result in completely unrealistic power system states, potentially harming training performance. 
\par 
In summary, the \ac{OPF}'s constraints, large-scale continuous action spaces, computational requirements, and lack of training data result in a challenging \ac{RL} problem. All this has to be considered in the environment design.

\par 
Considering the broadness of \ac{OPF} research and to limit the scope of this work, we will make some assumptions about the \ac{OPF} problems to solve, which also define the scope of this work.
We assume that the topology of the power system is fixed. We assume only continuous actuators, defining our environment's action space. For example, transformers as actuators are not considered. Further, we assume that one \ac{RL} agent is trained to solve a specific \ac{OPF} problem for a specific power system. Finally, we assume that the agent has full knowledge about the power system's state, as is the case in conventional \ac{OPF} solvers, \ie, no partial observability.   
All these are standard assumptions for learning the \ac{OPF} with \ac{RL}. Further, we explicitly focus only on the standard \ac{OPF} and not on related problems like, for example, real-time control of power systems \cite{fengStabilityConstrainedReinforcement2023} or multi-stage \acp{OPF} \cite{henryGymANMReinforcementLearning2021}.

\section{Environment Design Decisions}\label{sec:openDesign}
In the following, we will derive four categories of open design decisions and how they can be handled from the literature. The categories are training data, observation space, episode definition, and reward function. We will present the different options used in literature, explain the underlying idea and theory, and discuss the expected advantages and disadvantages. 
%

\subsection{Training Data}\label{sec:data}
One core question in all Machine Learning applications is the choice of training data. 
Regarding the \ac{OPF}, the goal is that the trained agent should be able to solve real-world \ac{OPF} problems and, therefore, work on real-world states of the electric system. 
However, an unsolved problem in energy research is that real-world data on power system states is very limited. And even if data is available, it is only possible to train on data from the past, while the application must work on future data in a highly dynamic system. Therefore, changing data distributions -- \ie changing load or generator behavior -- may result in performance degradation. 
\par 
The most straightforward training data option is to use time-series data of loads and generators, either realistic data collected from measurements or artificially created data. This environment design option \timeseries was used in references \cite{nieDeepReinforcementLearning2022, wolgast2023approximating,liuDeepReinforcementLearning2022,nieGeneralRealtimeOPF2019a}. 
The big advantage of time-series data is that it is closest to the actual data distribution in application. 
However, the main disadvantage of time-series data is that it is usually very limited, while deep learning methods require large amounts of data. Further, such data sets do not often contain special cases and extreme data points, \eg, holidays, extreme weather, etc., which bears the risk of failure in these uncommon cases.
Mathematically, the \timeseries option means we draw our power system states $s$ from a given finite dataset $\mathcal{D}$.
\begin{equation}
    s \sim \mathcal{D}
\end{equation}
\par 
The second option is to train on randomly sampled data, for example, by sampling all load and generator setpoints from a Normal distribution. 
This tackles both previously mentioned disadvantages of time-series data. Random data can be created in any quantity and does not favor common over uncommon states. Further, it can be used even without knowing the actual data distribution.
However, most randomly created data points will be completely unrealistic. Most distributions in power systems are highly coupled, \eg, wind turbine feed-in or load behavior. 
In this work, we will consider two random sampling options: a \uniformSampling training data distribution from references \cite{zhouDatadrivenMethodFast2020a, wooRealTimeOptimalPower2020, zhouDeepReinforcementLearning2022}
\begin{equation}
    s \sim \mathcal{U}(s_\minSub, s_\maxSub)
\end{equation}
and a \normalSampling  distribution around the mean \cite{yanRealTimeOptimalPower2020}
\begin{equation}
    s \sim \mathcal{N}(\mu, \sigma^2)
\end{equation}
with the data range $[s_\minSub, s_\maxSub]$, the mean $\mu$, and the variance $\sigma^2$.
Here, we assume that the required data range is known for uniform sampling and that the mean of the data is known for normal sampling.
The state variables are usually non-controllable active and reactive setpoints of loads and generators but can also be information about constraints or the current market situation, depending on the exact \ac{OPF} use case.  

\subsection{Observation Space}\label{sec:obs}
Another important question of \ac{RL} is which observations the agent receives to make its decision. 
The Markov property is fulfilled when the agent receives all the information required to make an optimal decision \cite{suttonReinforcementLearningSecond2018}.
From a power system perspective, only all active and reactive power values of all non-controllable loads and generators are required (including storages, shunts, etc.). 
Additionally, in market-based \acp{OPF}, we must consider all prices to enable optimal decisions.
System variables like voltages or line flows are not required because they can be computed from the node power values using power flow calculation.
\par 
Therefore, based on the Markov property, option \markov for the observation space is to use only the active/reactive power of all non-actuator units in the system, plus all market prices, if applicable. If the constraints change, we also need to provide that information to the agent. The \markov option is used in references \cite{yanRealTimeOptimalPower2020,zhouDatadrivenMethodFast2020a,nieDeepReinforcementLearning2022,wolgast2023approximating,liuDeepReinforcementLearning2022, zhenDesignTestsReinforcementlearningbased2021} from the \ac{RL}-\ac{OPF} literature. 
\begin{equation}
    \obs_\mathrm{Markov} \in \{ \activepower_\mathrm{load}, \reactivepower_\mathrm{load}, \activepower_\mathrm{gen}, \reactivepower_\mathrm{gen}, ...\}
\end{equation}
\par 
The redundant system variables are not used in the \markov option. However, most constraints in the \ac{OPF} concern exactly these system variables, \eg, the voltage band or maximum line loading. 
Therefore, these data may help with better decision-making, although being redundant from a theoretical standpoint. In consequence, design option \redundantObs adds the system variables to the observations of option \markovShort, namely voltages of all nodes in the system, loading of all lines and transformers in the system, and power flows from/to all external grids. This option is used in references \cite{wooRealTimeOptimalPower2020,zhouDeepReinforcementLearning2022,nieGeneralRealtimeOPF2019a}.
\begin{equation}
    \obs_\mathrm{Redundant} \in \obs_\mathrm{Markov} + \{\voltage_\mathrm{bus}, \loading_\mathrm{line}, \loading_\mathrm{trafo}, \loading_\mathrm{ext} \}
\end{equation}
To compute these system variables, we must calculate the power flow when a new data point is sampled in the environment. That requires initial actuator setpoints before the actual agent action. For that initial action handling, we consider two variants: the average possible action for all actuators
\begin{equation}
    \act_\mathrm{init} = (\act_\mathrm{max} - \act_\mathrm{min}) / 2
\end{equation}
or a random action:
\begin{equation}
    \act_\mathrm{init} \sim [\act_\minSub, \act_\maxSub]
\end{equation}
In the second variant, this action must also be added to the observation space for the agent to consider its influence on the observed state.
The potential advantage of option \redundantObsShort is better performance, especially regarding constraint satisfaction.
However, the required initial power flow calculation increases the computation time of the environment because two power flow calculations are required instead of one. Additionally, the bigger observation space may result in increased training time. 

\subsection{Episode Definition}\label{sec:episode}
In \ac{RL}, an \textit{episode} is the amount of steps until a termination criterion is reached. Usually, the agent aims to maximize the sum of rewards over an episode \cite{suttonReinforcementLearningSecond2018}. 
Two general options to define the \ac{RL} episodes to learn the \ac{OPF} are used in literature: the 1-step environment and the n-step environment. 
\par
Option \onestep means that every environment state is terminal, which starts a new episode. The agent receives an observation, performs a 1-shot action, and that action results in the final state, which is used for performance evaluation. This variant was chosen in references \cite{wolgast2023approximating,nieGeneralRealtimeOPF2019a,zhenDesignTestsReinforcementlearningbased2021}.
The advantage is that this 1-step episode simplifies training. Most \ac{RL} algorithms work by first predicting the expected sum of future rewards to then optimize for the actions that maximize that value. In the special case of a 1-step environment, the sum of rewards is simply the current reward. Therefore, the task of predicting becomes a simple supervised learning problem, without typical problems of n-step \ac{RL} like value overestimation \cite{hasseltDoubleQlearning2010}. 
The disadvantage, however, is that the agent cannot learn to correct its own mistakes, \eg, by observing a self-induced constraint violation and then performing an action to resolve that violation. This is not possible because the environment resets to a completely new state after each action. 
\par
In option \nstep, the agent can perform multiple actions to solve a given \ac{OPF} problem iteratively, which is more similar to conventional solvers. This sequential decision making is the normal case for \ac{RL} environments and used in \cite{yanRealTimeOptimalPower2020,zhouDatadrivenMethodFast2020a,nieDeepReinforcementLearning2022,wooRealTimeOptimalPower2020,zhouDeepReinforcementLearning2022,liuDeepReinforcementLearning2022}. 
The agent receives an initial observation, iteratively acts on these states, and receives the updated states as new observations. This can be repeated $n$ times, potentially improving the \ac{OPF} solution each time. 
The expected advantage is that the agent can learn to solve a given \ac{OPF} problem step-by-step, \eg, by correcting observed self-induced violations. 
However, as discussed before, the task of predicting future rewards is a more difficult learning task, which could slow down training or even decrease performance.  
Further, the agent will be trained with less diverse data because it interacts $n$ times with the same data point, resulting in $n$ times less data points during training if we train the same number of steps as option \onestepShort. 
\par 
Note that \nstep automatically includes the observation space variant \redundantObs because the agent requires the additional observations of voltage level and power flows to improve its actions. Only providing the \markov data to the agent does not work with \nstepShort because these values are independent of the agent's action, which means the agent will receive the same observation again and again.

\begin{table*}[t]
\centering
\footnotesize
\caption{\ac{RL}-\ac{OPF} literature overview including their environment design decisions.}
\begin{tabular}{lcccc}\toprule
    Reference                                   & Training Data & Observation Space  & Episode Definition  &  Reward Function  \\ \midrule 
    Zhou \etal \cite{zhouDeepReinforcementLearning2022}    & \uniformSampling & \redundantObs & \nstep & \replacement   \\ 
    Wolgast and Nieße \cite{wolgast2023approximating}             & \timeseries & \markov & \onestep & \summation \\
    Yan and Xu \cite{yanRealTimeOptimalPower2020}   & \normalSampling & \markov & \nstep & N.A.   \\ 
    Woo \etal \cite{wooRealTimeOptimalPower2020}   & \uniformSampling & \redundantObs & \nstep & \summation   \\ 
    Zhou \etal \cite{zhouDatadrivenMethodFast2020a}   & \uniformSampling & (\markov) & \nstep & \replacement   \\ 
    Liu \etal \cite{liuDeepReinforcementLearning2022}   & \timeseries & \markov & \nstep & \summation   \\ 
    Nie \etal \cite{nieGeneralRealtimeOPF2019a}   & (\timeseries) & \redundantObs & \onestep & \summation   \\ 
    Nie \etal \cite{nieDeepReinforcementLearning2022}   & \timeseries & \markov & \nstep & \summation   \\  
    Zhen \etal \cite{zhenDesignTestsReinforcementlearningbased2021}   & ? & \markov & \onestep & \replacement   \\ 
    \bottomrule
    \label{tab:literature}
\end{tabular}
\end{table*}

\subsection{Reward Function}\label{sec:reward}

The reward function is another core element of \ac{RL}. Therefore, we expect the reward design to greatly influence the agent's performance. 
When learning the \ac{OPF}, the agent has two goals: optimization of the objective function and constraint satisfaction. 
\par 
The objective function can be used as a negative reward (assuming a minimization problem). 
However, considering the constraints is not as straightforward because constraints are not part of the standard \ac{RL} framework. 
The common approach from the \ac{RL}-\ac{OPF} literature is to use penalties for invalid states in the reward \cite{khaloieReviewMachineLearning2024}. Two approaches are common, which we call \summation and \replacement in the following. 
\par 
The \summation method used in \cite{wolgast2023approximating,wooRealTimeOptimalPower2020,liuDeepReinforcementLearning2022,nieGeneralRealtimeOPF2019a,nieDeepReinforcementLearning2022} defines the reward as the sum of the objective function $\objectiveFct$ minus some penalty function $\penaltyFct$:
\begin{equation}
    R_\mathrm{sum} = - \objectiveFct - \penaltyFct
\end{equation}
with usually a linear penalty function $\penaltyFct$:
\begin{equation}
    \penaltyFct = \sum_{\constraint  \in \constraints} \penaltyFactor_{\constraint} \cdot \violation_{\constraint}
\end{equation}
with the violations $\violation$ and penalty factors $\penaltyFactor$ of all constraints $\constraints$. The expected advantage of \summationShort is that the agent can learn constraint satisfaction and optimization concurrently. However, the penalty factors need to be chosen well. If the penalties are too high, the agent will neglect optimization and focus on constraints only. If the penalties are too low, the agent might tolerate minor violations for better optimization if the optimum of the objective function is in the invalid state space.
\par 
The \replacement method used in \cite{zhouDatadrivenMethodFast2020a,zhouDeepReinforcementLearning2022, zhenDesignTestsReinforcementlearningbased2021} tackles that trade-off problem by only granting rewards for optimization if all constraints are satisfied:
\begin{equation}
    R_\mathrm{replace} = \begin{cases}
                            -\objectiveFct + \constant \mathrm{\;if\;valid} \\
                            -\penaltyFct \mathrm{\;else}
                         \end{cases}
\end{equation}
However, considering that the objective function $\objectiveFct$ can take negative values, we must add some constant value $\constant$ to the valid part of the reward function. This should ensure that all valid states yield higher rewards than all invalid states. Our work considers two general heuristics to choose $\constant$. In both, we sample \nInit random states with random actions and calculate their objective values. In variant 1, we set $\constant$ to the absolute worst-case sampled value. However, considering that the worst-case value bears the risk of an extremely high offset and a jump in the reward function, we also look into variant 2 of using the mean value instead. This way, we can also investigate the influence of that parameter choice. 
\par 
The expected advantage of the \replacementShort method is that the agent cannot tolerate constraint violations in any way. However, since the agent only receives sparse feedback on its optimization performance, training can be expected to be less efficient regarding optimization.    

\subsection{Summary}\label{sec:related}
In the previous sections, we identified four different aspects of environment design where the \ac{RL}-\ac{OPF} literature is strongly divided regarding their choice. 
While the observation space, reward function, and episode definition are unique to \ac{RL}, the training data distribution is also relevant for supervised learning. 

All four categories were also investigated by Reda \etal in their work \cite{redaLearningLocomoteUnderstanding2020}, which indicates that they are important for environment design on a general level, not only for the \ac{OPF}. 
\par 
Table \ref{tab:literature} categorizes all the previously mentioned references regarding their respective design decisions. Some of these literature design decisions do not perfectly fit our previous descriptions, which we indicated by putting them in brackets. Yan and Xu \cite{yanRealTimeOptimalPower2020}, did not use a reward function in the classical sense. Zhen \etal \cite{zhenDesignTestsReinforcementlearningbased2021} did not provide information about the training data distribution used. 
\par 
The overview demonstrates a high diversity regarding environment design. However, some design options are used significantly more often than others, \eg, the \summation reward or the \nstep episode definition. For a more comprehensive literature overview, refer to Khaloie \etal's recent review \cite{khaloieReviewMachineLearning2024}. 

\section{Environment Framework}\label{sec:framework}
For a systematic comparison of the presented design options, we need a general \ac{OPF} environment framework, which allows to define all kinds of \ac{OPF} problems as \ac{RL} environments. We build that framework on top of the \gymnasium\gymnasiumfootnote library, the de facto API standard for \ac{RL} environments. 
For the power system modeling and calculation, we use \pandapower\pandapowerfootnote and \simbench\simbenchfootnote, both broadly used in research and open-source. \pandapower provides power flow calculations and conventional \ac{OPF} calculations. The former are required to calculate the new state that results from an agent's action. The latter are required for performance testing the agents after training. \simbench provides benchmark power systems from all voltage levels and realistic time-series data, which can be used as training and testing data. 
\par 
\begin{figure}[ht]
    \centering
    \includegraphics[width=9.0cm]{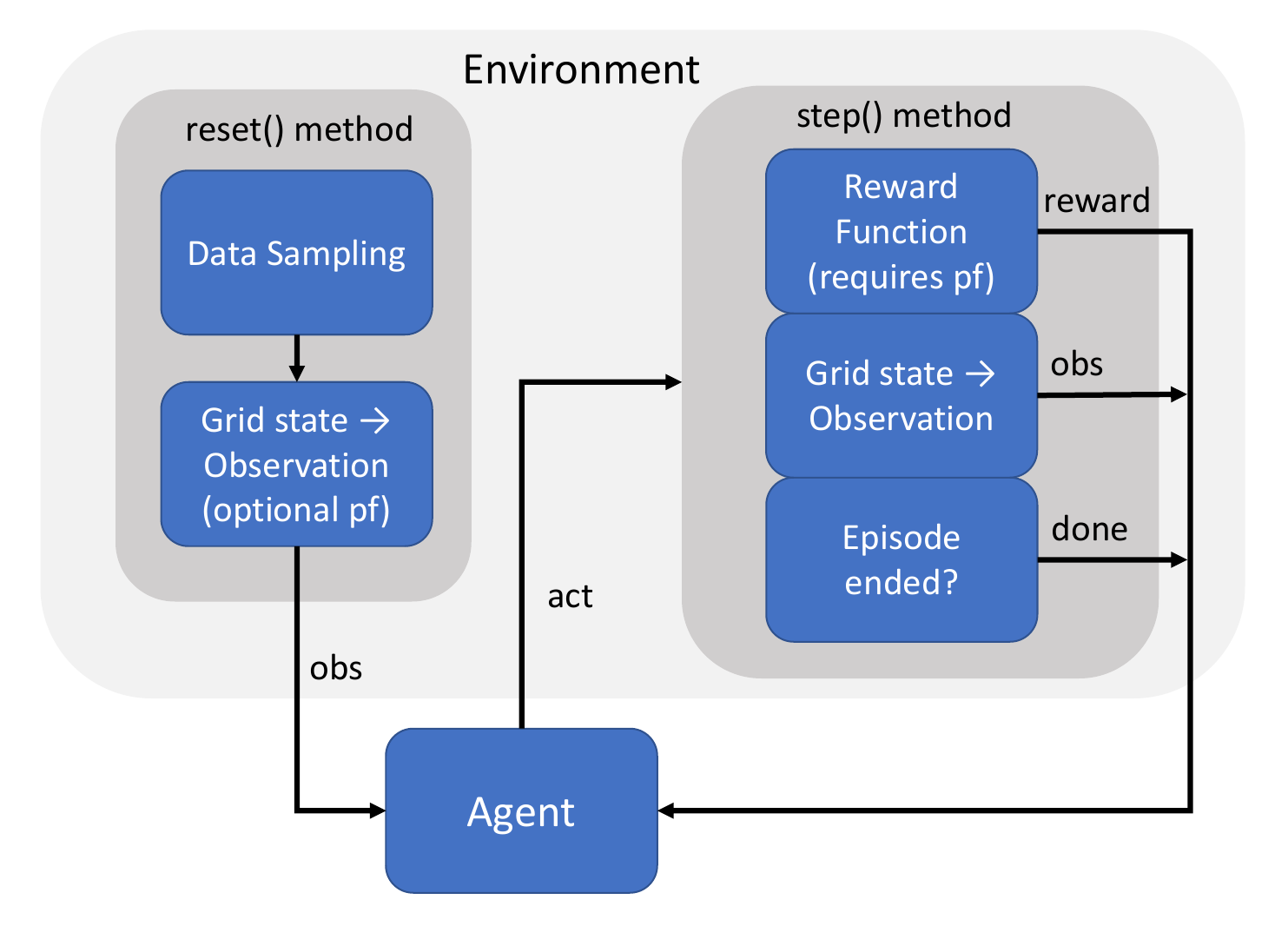}
    \caption{The procedure and API of the developed \ac{RL}-\ac{OPF} environment framework, following the \gymnasium API.}
    \label{fig:environment}
\end{figure}
The overall procedure of the framework is visualized in Figure \ref{fig:environment} and follows the \gymnasium API. 
The environment starts with a \reset method, which contains data sampling and the creation of the initial agent observation. Here, the power system is set to some random state for which the \ac{OPF} should be solved. That random state is sampled from some distribution, which is an open design decision discussed in section \ref{sec:data}. After sampling, the state of the power system is translated to the agent's observation, which should provide all information to compute optimal actions. The details of which observations should be provided is another open design decision discussed in section \ref{sec:obs}. 
Based on the observations, the agent performs an action. In the \ac{OPF}, these actions are usually continuous active and/or reactive power setpoints of units like generators, loads, or storages. 
The agent action is given to the environment's \step method, which applies the actions to the power system. Then, a power flow calculation is performed to compute the new state of the system, \eg, power flows and voltage values.    
Based on that new state, the reward and an updated observation are computed and given to the agent as learning feedback. The details of the reward function are a design decision discussed in section \ref{sec:reward}. Finally, it is determined if the episode ends, \ie, if the agent performs another \step or if the environment is \reset to a new state, as discussed in section \ref{sec:episode}.
Note that no data sampling in the \step method is performed, which means we are focusing on single time-step \ac{OPF} problems, in contrast to multi-stage \acp{OPF}.

\par 
The presented framework allows to define almost arbitrary \ac{OPF} problems and convert them to \ac{RL} environments. The objective function and the constraints are automatically extracted from the \pandapower model. Therefore, all \ac{OPF} problems possible in \pandapower can be converted to \ac{RL} environments. However, custom objectives and constraints are possible as well. The same is true for custom observations, data sets, etc. The framework is fully open-source and will continue to be developed.
    
\section{Analyzing RL Environment Design}\label{sec:experimentation}

This section defines the exact experimentation of our work. First, we will introduce the two \ac{OPF} problems to be converted to \ac{RL} environments for our experiments.
Further, we will define the default environment design to be used as baseline, the \ac{RL} algorithm and hyperparameter choice, as well as the experiment details and evaluation metrics. 

\subsection{Exemplary OPF Problems as RL Environment Instances}
The experiments regarding the choice of \ac{RL} environment design options are done for two different use cases in power systems where \ac{OPF} is relevant in the field: voltage control and economic dispatch. The simpler \voltagecontrol environment and the more complex \ecodis environment will be presented in the following, both following the previously described general framework.

\subsubsection{VoltageControl}
The first \ac{RL} environment is the VoltageControl problem, which can be formulated as \ac{OPF} problem \cite{sunReviewChallengesResearch2019, wolgastReactivePowerMarkets2022}. 
The objective is to minimize active power losses $\activepower_\mathrm{loss}$ in the system with minimal reactive power costs in state $s$:
\begin{equation}
    \objectiveFct = \Min \activepower_\mathrm{loss} \cdot \price^\activepower + \sum_{\generator \in \generators} \reactivepower^2_\generator \cdot \price_\generator^\reactivepower
\end{equation}
with the active power price $\price^\activepower$, the reactive power price $\price_\generator^\reactivepower$ and reactive power $\reactivepower$ of the generators $\generators$. We assume a quadratic reactive power price, according to \cite{samimiEconomicenvironmentalActiveReactive2015}. Since we consider reactive power prices in the optimization, the VoltageControl problem is equivalent to a reactive power market problem, also usually formulated as \ac{OPF} \cite{wolgastReactivePowerMarkets2022}.
\par 
The constraints are the voltage band of all buses $\buses$,  
\begin{equation}\label{eq:voltageBand}
    \voltage_\minSub \leq \voltage_\bus \leq \voltage_\maxSub  \;\forall \; \bus  \;\in \; \buses
\end{equation}
maximum line loading of all lines $\lines$,
\begin{equation}\label{eq:loading}
    \loading_\lineSingle \leq 100\%  \; \forall  \;\lineSingle \; \in \; \lines
\end{equation}
maximum power exchange with the external grid (\ie slack bus),
\begin{equation}
    -\reactivepower_\maxSub \leq \reactivepower_\grid \leq \reactivepower_\maxSub
\end{equation}
and the maximum reactive power of the generators $\generators$
\begin{equation}
    -\maxReactive \leq \reactivepower_\generator \leq \maxReactive  \forallGens
\end{equation}
\begin{equation}
    \maxReactive = \sqrt{\apparentPower_{\maxSub, \generator}^2 - \activepower_\generator^2} \forallGens,
\end{equation}
which is calculated from the current fixed active power setpoint $\activepower$ and the maximum apparent power $\apparentPower_{\maxSub}$. Note that the environment does not need to contain the power balance equations because they are automatically handled in the power flow calculation used to calculate the next environment state. 
\par 
The degrees of freedom / actions $\act$ are the reactive power setpoints $\reactivepower$ of all generators $\generators$. Actions are translated to setpoints as follows:
\begin{equation}
    \reactivepower_\generator = \act_\generator \cdot \maxReactive \; \mathrm{with} \; \act \in [-1, 1]
\end{equation}
\par 
Our observations to fulfill the Markov property are the active and reactive power of all loads, the reactive power prices of all generators, and the active power setpoints of all generators. Optionally, we can add line loadings, voltage values, and the power exchange with the external grid (see \redundantObs observation in \ref{sec:obs})
\par 
As power system, we use the \texttt{1-LV-urban6--0-sw} \simbench system with 59 buses, 111 loads, and five generators. 
The resulting VoltageControl environment has five continuous actions and  233 observations (352 with \redundantObs). 
\par
\begin{figure}[ht]
    \centering
    \includegraphics[width=8.5cm]{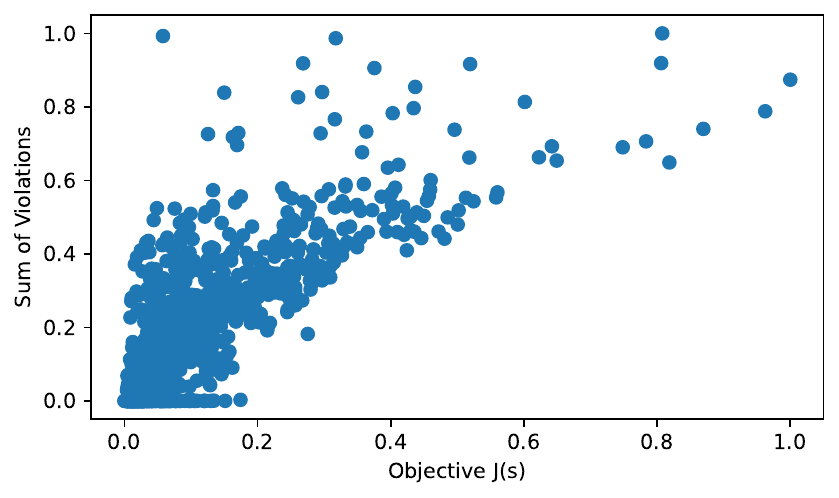}
    \caption{VoltageControl: Scatter plot of normalized objective values and sum of violations.}
    \label{fig:qmarket}
\end{figure}

To analyze the behavior and complexity of the VoltageControl environment, Figure \ref{fig:qmarket} shows a scatter plot of the normalized objective values and summed violations when performing $\nScatter$ random actions in randomly sampled states from the \simbench data set. For both metrics, lower means better. The plot shows a strong positive Pearson correlation of $+0.78$ of objective values and violations. This means an action that creates a valid solution usually has good optimization performance, and vice versa. Therefore, we expect this positive correlation to simplify training. 
Because of the rather small action space and the positive violation-objective correlation, the VoltageControl environment represents a rather simple \ac{RL} problem. However, we observe some extreme negative outliers regarding objective function, which could pose a problem for learning.

\subsubsection{EcoDispatch}\label{sec:eco}
The second environment EcoDispatch represents an Economic Dispatch problem. 
The objective is to minimize active power costs:
\begin{equation}
    \objectiveFct = \Min \sum_{\generator \in \generators} \activepower_\generator \cdot \price^{\activepower}_\generator
\end{equation}
The constraints are again the voltage band (equation \ref{eq:voltageBand}) and the line loading (equation \ref{eq:loading}). Additionally, the active power exchange with the external system is restricted in a way that the agent must create power balance with the local generators only: 
\begin{equation}
    -\reactivepower_\maxSub \leq \reactivepower_\grid \leq 0
\end{equation}
The active power setpoints of the generators have a fixed boundary here, in contrast to the previous environment:
\begin{equation}
    \activepower_\generator \leq \activepower^\maxSub_\generator \forallGens.
\end{equation}
The actions $\act$ are the active power setpoints $\activepower$ of all generators $\generators$. Actions are translated to setpoints as follows:
\begin{equation}
    \activepower_\generator = \act_\generator \cdot \activepower^\maxSub_\generator \; \mathrm{with} \; \act \in [0, 1]
\end{equation}
The Markov observations are active and reactive power setpoints of all loads and the active power prices of all generators. 
\par 

As power system, we use the \texttt{1-HV-urban--0-sw} \simbench system with 372 buses, 79 loads, and 42 generators. 
The resulting EcoDispatch environment has 42 continuous actions and 201 observations (691 with \redundantObs). 
\par
\begin{figure}[ht]
    \centering
    \includegraphics[width=8.5cm]{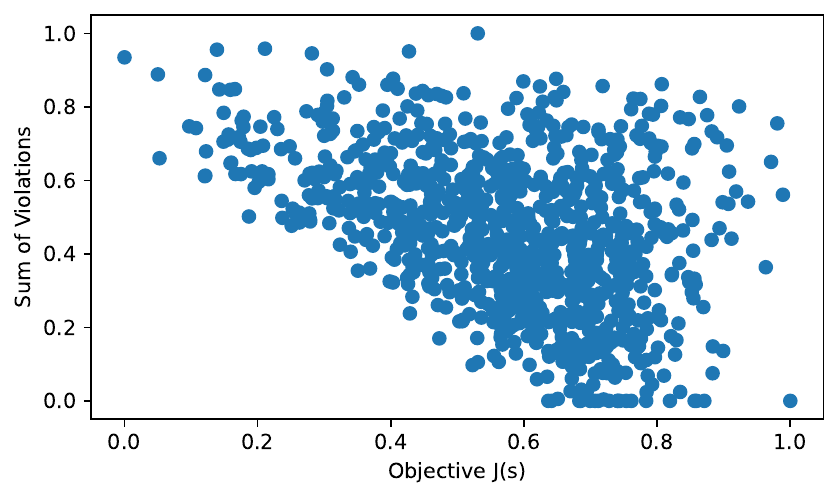}
    \caption{\ecodis: Scatter plot of normalized objective values and sum of violations.}
    \label{fig:ecodis}
\end{figure}

Figure \ref{fig:ecodis} shows the scatter plot of violations and objective values resulting from random actions in random states again. In contrast to the previous case, we now observe a negative correlation of $-0.42$. Therefore, a valid solution tends to have worse optimization performance, and vice versa. That inherent trade-off of optimization and constraint satisfaction performance can be expected to complicate training. Additionally, the environment has a 42-dimensional action space, which is very large for a \ac{RL} problem. 
\par 
Altogether, the VoltageControl environment and the EcoDispatch environment were designed to represent two levels of difficulty of the \ac{RL}-\ac{OPF} problem. 

\subsection{Default Environment Settings}\label{sec:baseline}
To investigate the performance impacts of the environment design decisions, we need to perform experiments in which only the respective design decisions get altered, while all other settings remain unchanged. Therefore, we define default designs. For this work, we use \timeseries sampling for training data distribution, the \markov observation space, the \onestep episode definition, and the \summation reward function design with penalty factors of 500 for VoltageControl and 10000 for EcoDispatch as default. 
\par 
To test the agents' performance during and after training, we sample from a separate test data set, which is not used in training. The test data set was created by separating 20\% of the \simbench data before training, resulting in 6720 test samples and 28416 training samples (only relevant for option \timeseries). This way, even agents trained with the \timeseries training data distribution never observe these data during training, ensuring transferability to unseen data. The test sets are exactly the same for all performed experiments for better comparability of the results.

\subsection{RL Algorithm and Hyperparameters}
This work focuses on the environment design for \ac{OPF} problems. The goal is that the resulting environment design works for standard \ac{RL} algorithms. Accordingly, we pick \ac{DDPG} \cite{lillicrapContinuousControlDeep2019} for our experiments without restricting the general applicability of our approach. \ac{DDPG} is the most used \ac{RL} algorithm in \ac{RL}-\ac{OPF} literature, \eg, used in \cite{yanRealTimeOptimalPower2020,nieDeepReinforcementLearning2022,liuDeepReinforcementLearning2022,nieGeneralRealtimeOPF2019a}. Additionally, it is an off-policy algorithm, making it sample-efficient and advantageous for computationally expensive environments like \ac{OPF} problems. The hyperparameter choices can be found in \ref{app:hyperparameters}.

\begin{figure*}[ht]
    \centering
    \includegraphics[width=15cm]{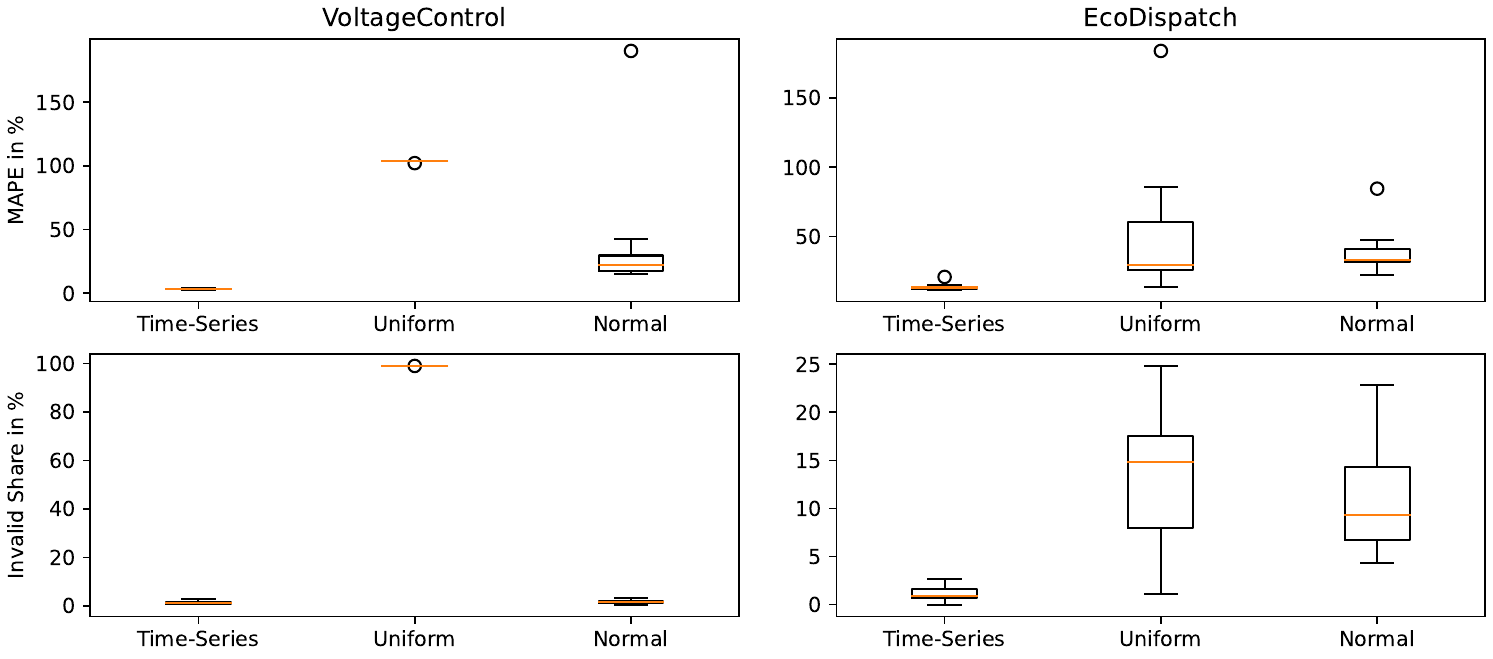}
    \caption{Training Data - Comparison of design options regarding optimization \acs{MAPE} (first row), share of invalid solutions (second row), and variance in both the \voltagecontrol environment and the \ecodis environment (arranged in columns).}
    \label{fig:data}
\end{figure*}
\subsection{Experiments and Evaluation Metrics}
To consider the stochasticity of \ac{RL} experiments, we repeat every training run ten times \cite{hendersonDeepReinforcementLearning2018}. This also allows us to evaluate the variance of results over the training runs, which can serve as a metric to evaluate design decisions regarding their robustness. We train for one million steps in the VoltageControl and two million steps in the EcoDispatch environment. 

\par 
We aim to evaluate the different environment design decisions based on multiple metrics. To evaluate the capability of the \ac{RL} agent to solve the \ac{OPF} optimization problem, we compute the \acf{MAPE} relative to the optimal solution from a conventional solver. To ensure that the agent cannot buy better optimization performance by creating invalid system states, we use only valid samples for the \ac{MAPE} computation: 
\begin{equation}
    MAPE = \left(\frac{1}{N_\valid} \sum_{N_\valid} \frac{|\objective - \objective^*|}{\objective^*}\right) \cdot 100\%
\end{equation}
with the number of samples $N$ and optimal objective value $J^*$. To evaluate the agents' capability for constraint satisfaction, we compute the share of invalid solutions in the test data set:
\begin{equation}
    \left(1 - \frac{N_\valid}{N}\right) \cdot 100\%
\end{equation}
Both metrics need to be minimized for improved performance. Additionally, we will evaluate the design decisions based on computational effort and variance of results as soft metrics. In both cases, less is better to ensure quick training and reproducible performance.   

\begin{figure*}[ht]
    \centering
    \includegraphics[width=15cm]{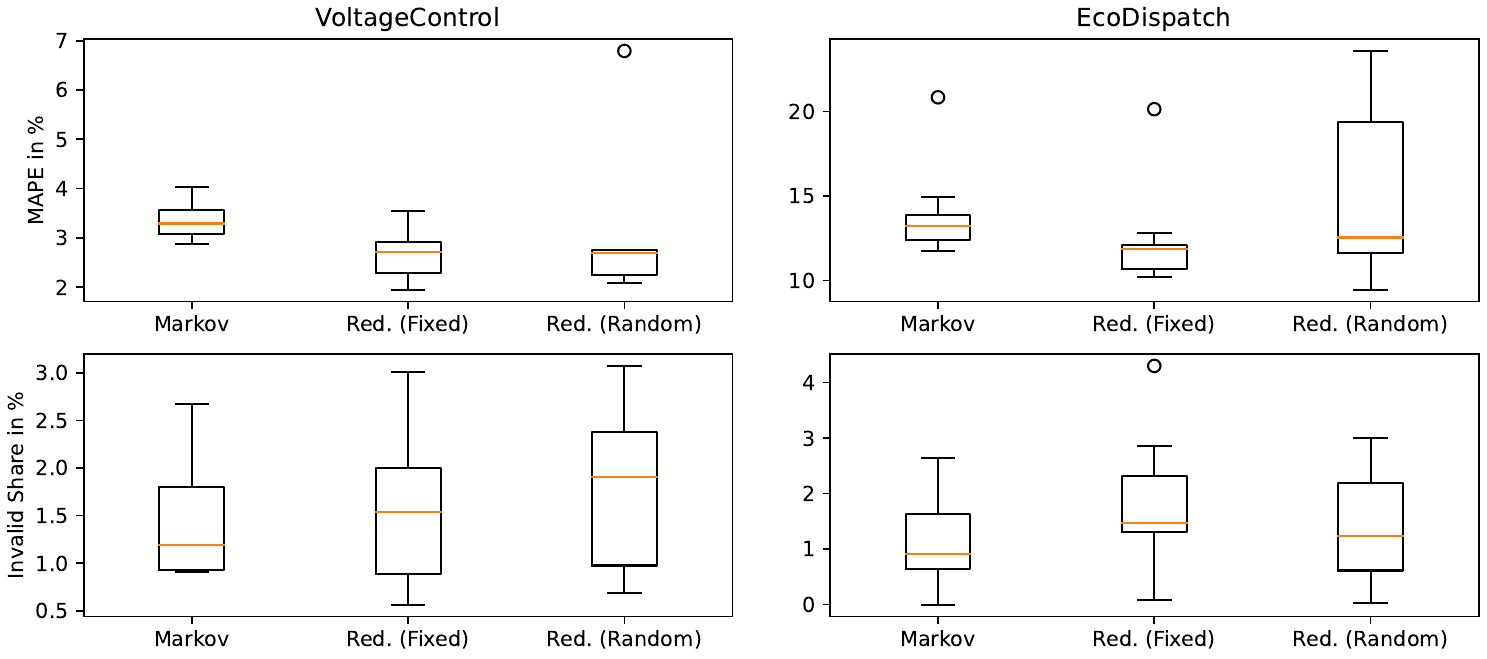}
    \caption{Observation Space - Comparison of design options regarding optimization \ac{MAPE} (first row), share of invalid solutions (second row), and variance in both the \voltagecontrol environment and the \ecodis environment (arranged in columns).}
    \label{fig:obs}
\end{figure*}

\section{Evaluation}\label{sec:results}

The main objective of this work is to compare and analyze different design decisions in \ac{RL}-\ac{OPF} environments regarding their influence on training performance. In the following, we present the training results of the four categories presented in section \ref{sec:openDesign}, discuss the influence of the respective options on the performance metrics, and draw recommendations for \ac{RL}-\ac{OPF} environment design.

\subsection{Data Sampling Distribution}\label{sec:dataRes}
Figure \ref{fig:data} shows the results regarding the training data distribution for both environments and both metrics. For the \normalSampling sampling method, we use a standard deviation of 0.3 relative to the respective data range of the unit.
The first row shows the \ac{MAPE} metric, the second row shows the share of invalid system states that resulted from the trained agent's actions, and the columns represent the two environments. We use boxplots to visualize the median and distribution over all ten experiment runs. The leftmost boxplot is always the baseline environment setting described in section \ref{sec:baseline}. We will use this data visualization for all the following design decision experiments.
\par 
\paragraph{Results}
The results show that the \timeseries option drastically outperforms the random sampling options regarding both metrics and in both environments. The \ac{MAPE} and invalid share are significantly and reproducibly lower. In some cases, the \normalSampling sampling is competitive, \eg, regarding constraint satisfaction in VoltageControl (1.57\% vs. 1.19\%). However, \uniformSampling sampling is consistently outperformed. In the VoltageControl environment, it even almost completely failed to produce valid solutions. Additionally, we observe more variance in both random sampling experiments. 
\par 
\paragraph{Discussion}
The results confirm the hypothesis that training on randomly sampled data can result in bad performance if tested on realistic data. This can be explained by having too many unrealistic system states in the training data. For example, we investigated the bad performance of \uniformSampling and found that above 99\% of these states are not solvable by the conventional solver, although being sampled from the \simbench data range. That explains why the \ac{RL} agents almost completely fail to learn successful constraint satisfaction.  However, the results also demonstrate that even little knowledge about the test data distribution can significantly improve the performance of random sampling. In the \normalSampling option, the respective data mean is used for sampling. That additional knowledge seems to improve performance drastically, compared to the \uniformSampling sampling, which covers the whole possible range uniformly.
\paragraph{Recommendation}
Due to the clear results, we recommend using realistic time-series data for training if available. 
However, there is a high potential of overfitting to the usually very limited data and failure in unusual scenarios, which should always be accounted for.  More experiments in future work will clarify whether randomly sampled data can be useful in other use cases.   
Potentially, the solution could be artificially created large time-series data sets combined with additional random noise. 

\subsection{Observation Space}\label{sec:obsRes}
Figure \ref{fig:obs} shows the observation space results for both environments and metrics. We investigated the \redundantObs option in two variants regarding the default starting action, as described in \ref{sec:obs}. In variant \textit{fixed}, the environment always starts with the average of all actions. In variant \textit{random}, a random action is sampled, applied, and added to the observation. In both cases, a power flow calculation is performed afterward to generate the additional observations. We consider these different variants to differentiate between the influence of the actual design decision and that of its implementation. 
\par 
\subsubsection{Results}
The results show that the \redundantObs observation variant outperforms the \markov variant regarding optimization, but underperforms regarding constraint satisfaction, which is the opposite of the expectations formulated in section \ref{sec:obs}. Further, we observe a slightly higher variance of the \redundantObs results of both variants. 
Additionally, the \textit{fixed} variant of \redundantObs seems to slightly outperform the \textit{random} variant.
However, all the mentioned differences are quite small, especially compared to the training data results in the previous section.
\par 
\paragraph{Discussion} 
Overall, adding redundant observations does not seem to improve performance in a meaningful way. It can even harm constraint satisfaction, which happened in both presented environments. However, the \redundantObs option requires a second power flow calculation, which slows down training. Since the power flow calculation is computationally very heavy, it increased training time by about 30\% in both environments.\footnote{We do not provide exact numbers regarding training time because we could not ensure the exact same hardware for all experiments. Additionally, the availability of GPUs plays a big role, which we also could not fully control during our experiments. Therefore, these are only rough numbers to indicate the general magnitude.}
\paragraph{Recommendation} 
Due to the unclear advantages of redundant observations and the computationally heavy additional power flow calculation, we recommend using the \markov observation space option. This recommendation is supported by literature where the \markov observation space is the favored solution as well (see Table. \ref{tab:literature}). 
However, exceptions are possible in special cases, \eg, if the power flow calculation is performed anyway or in the case of partial observability. 

\subsection{Episode Definition}\label{sec:episodeRes}
\begin{figure*}[ht]
    \centering
    \includegraphics[width=15cm]{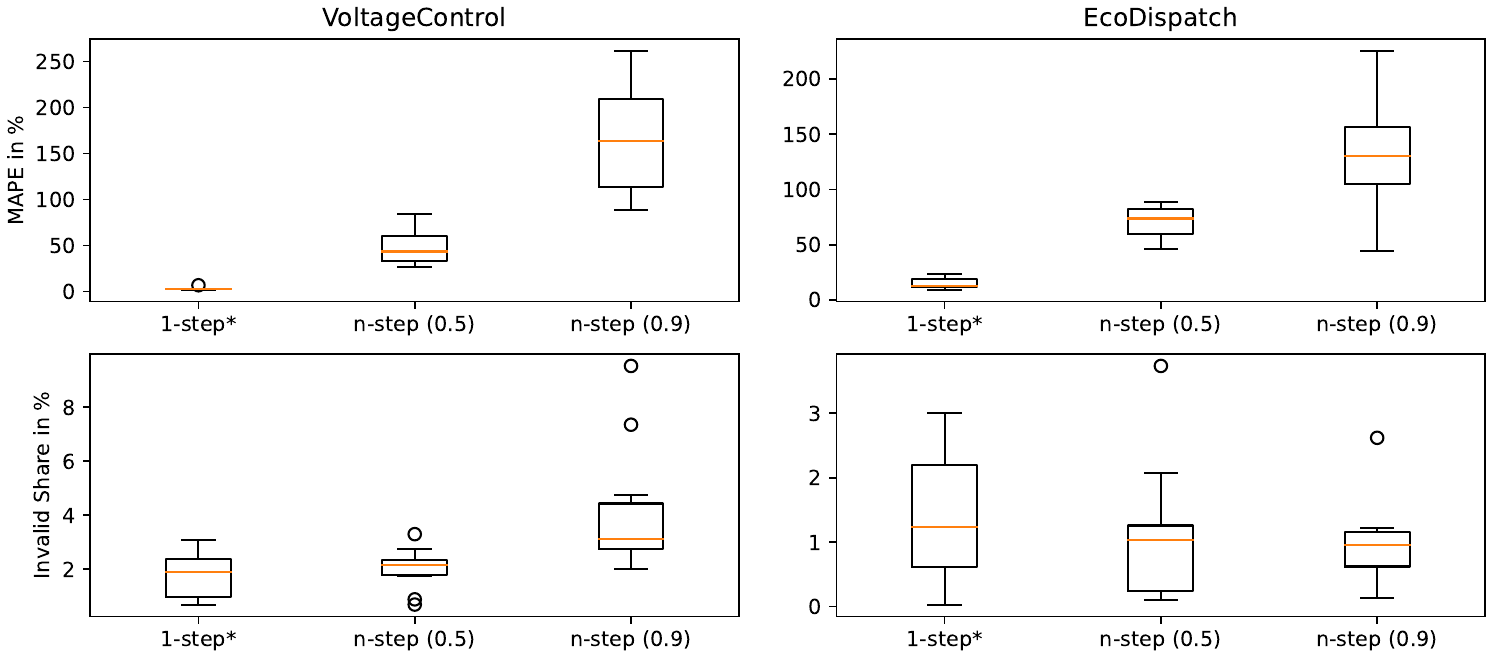}
    \caption{Episode Definition - Comparison of design options regarding optimization \ac{MAPE} (first row), share of invalid solutions (second row), and variance in both the \voltagecontrol environment and the \ecodis environment (arranged in columns). *Note: The baseline here is exceptionally the \redundantObs option with \textit{random} initialization.}
    \label{fig:episode}
\end{figure*}
Figure \ref{fig:episode} shows the results regarding the episode definition. Here, we use two different variants of the \nstep option. We performed the training for two different choices of the hyperparameter $\gamma$, which controls how greedy the agent maximizes its reward. We chose $\gamma=0.5$ for a short-sighted and $0.9$ for a more myopic agent. The episode length $n$ was chosen as five in both cases.  
Note that the \nstep option automatically includes the \redundantObs observation space with \textit{random} action initialization. Therefore, we exceptionally use \redundantObs as a baseline to ensure that the episode definition is the only influence factor for training performance. 
\paragraph{Results}
With one exception, the \onestep environment outperforms the \nstep variants, and the short-sighted \nstep variant outperforms the myopic variant significantly.
However, in the EcoDispatch environment, the $\gamma=0.9$ variant slightly outperforms the other two regarding constraint satisfaction (1-step: 1.24\%, \nstep 0.5: 1.03\%, \nstep 0.9: 0.96\% median invalid solutions). Additionally, the variance is lower here.  
\paragraph{Discussion}

Overall, the results indicate that rather short-sighted behavior is beneficial for \ac{RL}-\ac{OPF}. Not only does the \onestep environment win the overall comparison, the short-sighted \nstep variant comes in second. However, the slightly better constraint satisfaction of the myopic variant indicates that the \nstep environment can improve constraint satisfactions in some cases, as suspected in section \ref{sec:episode}. This comes to fruition in the EcoDispatch environment with its difficult trade-off of constraint satisfaction and optimization (see section \ref{sec:eco}). However, it requires the additional costly power flow calculation, because it includes the \redundantObs observation.
Overall, the additional complexity of a \nstep learning problem seems to outweigh the advantages of performing multiple trials. 
\paragraph{Recommendation}
Overall, we recommend to start with the \onestep environment definition. That is currently done by only the minority of publications (see Table \ref{tab:literature}). The \nstep option can be tested if constraint satisfaction is not satisfactory. It also seems to be the natural choice for multi-stage \ac{OPF} problems over multiple time steps, which we do not consider here. 

\subsection{Reward Function}\label{sec:rewardRes}
\begin{figure*}[ht]
    \centering
    \includegraphics[width=15cm]{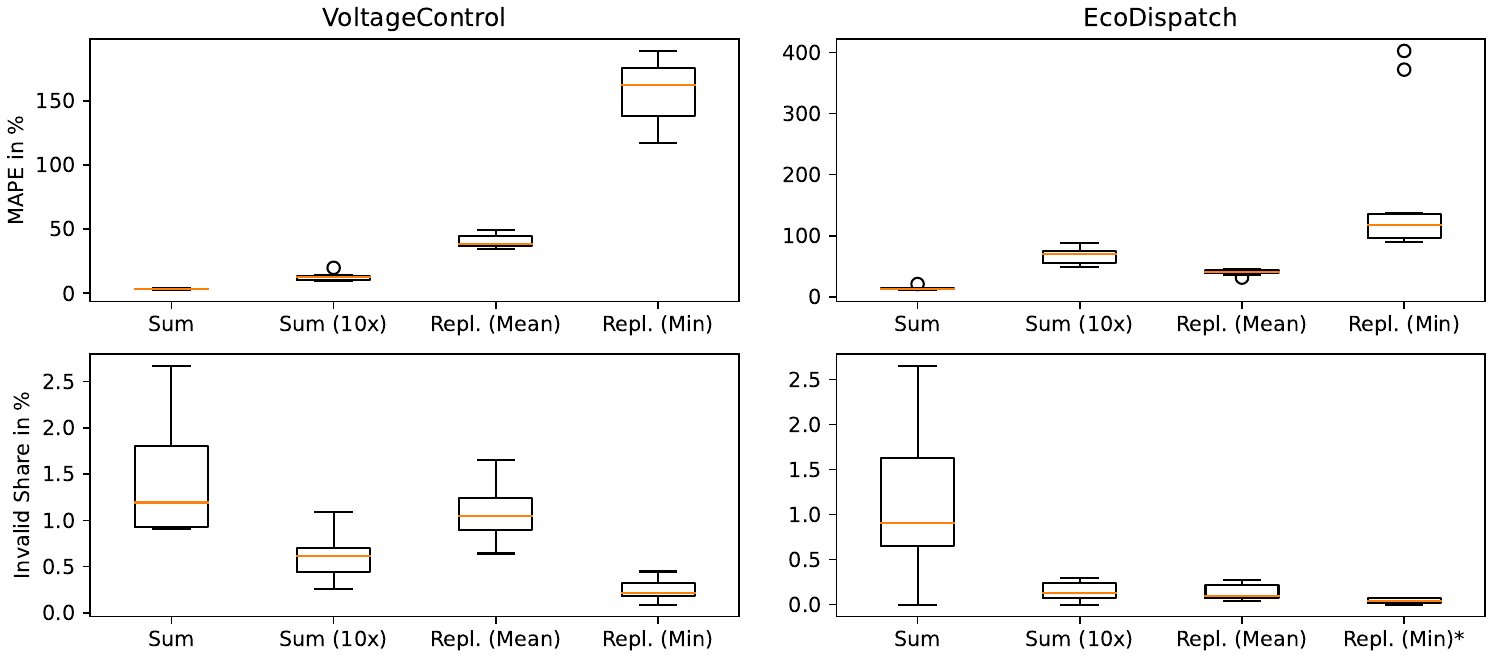}
    \caption{Reward Function - Comparison of design options regarding optimization \ac{MAPE} (first row), share of invalid solutions (second row), and variance in both the \voltagecontrol environment and the \ecodis environment (arranged in columns). *Note: We removed two outliers of 26\% and 69\% in the bottom right subplot of the \replacement (\textit{Min}) variant because their extreme magnitudes resulted in losing too much visual information.}
    \label{fig:reward}
\end{figure*}
Figure \ref{fig:reward} shows the results of the reward function choice. To investigate the influence of the penalty factors and reward offset $\constant$, we run experiments with multiple variants again. For the \summation method, we additionally investigate a penalty factor ten times higher than the baseline (variant \textit{10x}). For the replacement method, we choose the offset $\constant$ once as the mean and once as the worst-case reward, as described in section \ref{sec:reward}.
Preliminary experiments showed that the penalty factor of the replacement method is not as important as in the summation method because it does not compete with the objective part of the reward function. Therefore, we do not investigate multiple variants here. 
\paragraph{Results}
Overall, both reward options show a trade-off of constraint satisfaction vs. optimization in both environments. The variant with better constraint satisfaction also has a worse \ac{MAPE}, and vice versa. Further, the low penalty \summation method outperforms the other approaches regarding \ac{MAPE} but performs worst regarding constraint satisfaction in both environments. The \replacement (\textit{Min}) variant, on the other hand, performs worst regarding optimization and best regarding constraints satisfaction, however, also showing two extreme outliers. 
\par 
Apart from that, the results are very different for the two environments. In the VoltageControl environment, the high penalty \summation variant outperforms the \replacement (\textit{Mean}) variant regarding both metrics. In the EcoDispatch environment, constraint satisfaction is almost identical (\textit{10x}: 0.126\%, \textit{Mean}: 0.097\%), while the \replacement (\textit{Mean}) variant achieves significantly better optimization and less variance compared to high penalty \summation. 
\paragraph{Discussion}
Most results were already hypothesized in section \ref{sec:reward}. A too low penalty in the \summation method neglects constraint satisfaction. The very conservative choice of the offset $\constant$ in \replacement (\textit{Min}) results in very good constraint satisfaction but also neglects optimization and potentially destabilizes training, which happened in two out of ten training runs. 
\par 
The more ambiguous results can be explained by looking at the environment characteristics visualized in Figures \ref{fig:qmarket} and \ref{fig:ecodis}. The \summation method results in a concurrent improvement of constraint satisfaction and optimization, which benefits from the observed correlation of these two traits. However, in the EcoDispatch environment, \summation will result in a trade-off between constraint satisfaction and optimization, and the agent will prioritize the reward part that is more heavily weighted and sacrifice the other. The \replacement method prevents that trade-off by incentivizing first constraint satisfaction, then optimization, similar to curriculum learning. However, In the VoltageControl environment, that exact trait results in a performance decrease due to a sparser reward, as hypothesized in section \ref{sec:reward}. 
\paragraph{Recommendation}
The results show that the reward function needs to be chosen depending on the characteristics of the \ac{OPF} problem to solve and based on the importance of constraint satisfaction. The experiments indicate that in problems with conflicting objectives, the \replacement method prevents the agent from sacrificing constraint satisfaction for optimization and vice versa. If that is not the case, the \summation method enables a denser reward and better optimization performance. Additionally, its penalty factor allows for more fine-grained control of the constraint satisfaction importance. Therefore, it is also the better choice if smaller constraint violations can be tolerated. Overall, we recommend always investigating both but starting with the \summation method, which is also the favored approach in literature (see Table \ref{tab:literature}).
\par 
However, it remains unclear whether penalties are even the right approach to learn constraint satisfaction. The research area of Safe \ac{RL} is a promising alternative here and was successfully applied to similar problems already \cite{ceustersSafeReinforcementLearning2023}. We leave that question to future work.

\section{Conclusion}\label{sec:conclusion}
\paragraph{Contributions}
This work identified four categories of design decisions for \ac{RL}-\ac{OPF} environments, namely regarding training data, observation space, episode definition, and reward function. We systematically analyzed the respective impact of overall 13 variants on training performance in two \ac{OPF} environments. 
\par 
The overall message of our work is that environment design matters for the \ac{RL}-\ac{OPF}. It is an important step in solving \ac{OPF} problems with \ac{RL} algorithms, as been shown for other application areas by Reda \etal \cite{redaLearningLocomoteUnderstanding2020} and been hypothesized at the beginning of this contribution. Applied to the \ac{RL}-\ac{OPF}, it can have a significant impact on optimization performance (see section \ref{sec:dataRes}), constraint satisfaction (\ref{sec:rewardRes}), training time (\ref{sec:obsRes}), or simplicity of the learning task (\ref{sec:episodeRes}).
\par 
From our results, we derived some first recommendations for other researchers and practitioners on defining \ac{RL}-\ac{OPF} environments. 
The results in \ref{sec:dataRes} indicated that realistic time-series data should be used for training and cannot be replaced by naively sampling random data. Second, we could not show any significant advantages of using redundant observations like voltage values or power flows (\ref{sec:obsRes}). Therefore, since it requires an additional power flow calculation, we hypothesize it to be unnecessary and harmful regarding training time. Third, section \ref{sec:episodeRes} indicated that a \onestep episode is favorable over an \nstep episode. However, the \nstep variant slightly outperformed the \onestep environment in one case, making the results slightly ambiguous and requiring more research effort. 
Finally, section \ref{sec:rewardRes} shows that the reward function choice is highly problem-specific, depending on the exact \ac{OPF} problem and the weighting of constraint satisfaction vs. optimization performance. 
\par 
In summary, we found some design decisions that we expect to be transferable to other \ac{OPF} environments, while others seem to be more ambiguous and problem-specific. Therefore, it is advisable to perform systematic design experiments as we showed here for all newly created \ac{OPF} environments. Our results can provide a good starting point. Also, our environment framework, with all its implemented design options, will be open-sourced. Besides its application in environment design, it can also serve as a benchmark framework for future \ac{RL}-\ac{OPF} advances.
\paragraph{Future Work}
Considering some ambiguities in the results, more experiments in more different \ac{OPF} variants and different power systems are required to investigate if the derived recommendations are generally applicable or if the environment design must be chosen problem-dependent. Large-scale experiments will allow for very general recommendations on the design of \ac{RL}-\ac{OPF} environments. Further, similar experiments for different algorithm-hyperparameter combinations are required to investigate potential interconnections of the algorithm choice and the environment design.   

\section*{Acknowledgement}
Many thanks to Rico Schrage for his valuable feedback on this manuscript and to Fenno Boomgaarden for his literature research on reward function design in RL-OPF research.

\appendix

\section{Hyperparameter Choices}\label{app:hyperparameters}
Table \ref{tab:hyperparams} shows the \ac{DDPG} hyperparameters used for this work. Note that we used a bigger neural network for the EcoDispatch environment because of its greater complexity.

\begin{table}[ht]
\footnotesize
\centering
\caption{Hyperparameters for all experiments.}
\begin{tabular}{ll} \toprule
Hyperparameter          & Value   \\ \midrule
Memory size             & 1,000,000        \\
Batch size              & 1024           \\
Actor learning rate     & 0.0001         \\
Critic learning rate    & 0.0005         \\
Neurons/layer           & \voltagecontrol: $3 \times 256$      \\
                        & \ecodis: $6 \times 256$       \\
Optimizer               & Adam     \\
Noise std               & 0.1        \\ \bottomrule 
\label{tab:hyperparams}
\end{tabular}
\end{table}

 \bibliographystyle{elsarticle-num} 
 \bibliography{cas-refs}





\end{document}